\documentclass[lettersize,journal]{IEEEtran}
\usepackage{algorithm}
\usepackage{bbding,pifont}
\usepackage{amsthm,amsmath,amssymb}
\usepackage{algorithmicx}%
\usepackage{algpseudocode}%
\usepackage{graphicx}
\usepackage{subfig}
\usepackage{siunitx}
\usepackage[colorlinks,linkcolor=red]{hyperref}
\usepackage{color}
\usepackage[noadjust]{cite}
\usepackage{xcolor}
\usepackage{pifont}

\begin{document}
\title{Locating and Mitigating Gradient Conflicts in Point Cloud Domain Adaptation via Saliency Map Skewness}

\author{Jiaqi~Tang, ~Yinsong~Xu,~and~Qingchao~Chen\textsuperscript{\ding{41}}% <-this % stops a space
\thanks{This work was supported by the grant from the National Natural Science Foundation of China (62201014), Clinical Medicine Plus X - Young Scholars Project of Peking University, the Fundamental Research Funds for the Central Universities.}
\thanks{Jiaqi Tang and Qingchao Chen are with the National Institute of Health Data Science, Peking University, Beijing, 100191, China. They are also affiliated with the Institute of Medical Technology, Peking University, and the State Key Laboratory of General Artificial Intelligence, Peking University, Beijing, China. (e-mail: jiaqi.tang818@gmail.com, qingchao.chen@pku.edu.cn)}
\thanks{Yinsong Xu is with the School of Artificial Intelligence,
Beijing University of Posts and Telecommunications, Beijing 100876, China. He is also affiliated with the National Institute of Health Data Science, Peking University. (e-mail: xuyinsong@bupt.edu.cn)}
\thanks{\ding{41} Corresponding Author: Qingchao Chen}
% <-this % stops a space
% <-this % stops a space
% \thanks{Manuscript received April 19, 2005; revised August 26, 2015.}
}

% \markboth{Journal of \LaTeX\ Class Files,~Vol.~14, No.~8, August~2015}%
% {Shell \MakeLowercase{\textit{et al.}}: Bare Demo of IEEEtran.cls for IEEE Journals}



\maketitle

\begin{abstract}
Object classification models utilizing point cloud data are fundamental for 3D media understanding, yet they often struggle with unseen or out-of-distribution (OOD) scenarios. Existing point cloud unsupervised domain adaptation (UDA) methods typically employ a multi-task learning (MTL) framework that combines primary classification tasks with auxiliary self-supervision tasks to bridge the gap between cross-domain feature distributions. However, our further experiments demonstrate that  \textit{not all gradients from self-supervision tasks are beneficial and some may negatively impact the classification performance.}
In this paper, we propose a novel solution, termed Saliency Map-based Data Sampling Block (SM-DSB), to mitigate these gradient conflicts. Specifically, our method designs a new scoring mechanism based on the skewness of 3D saliency maps to estimate gradient conflicts without requiring target labels. Leveraging this, we develop a sample selection strategy that dynamically filters out samples whose self-supervision gradients are not beneficial for the classification. Our approach is scalable, introducing modest computational overhead, and can be integrated into all the point cloud UDA MTL frameworks.
Extensive evaluations demonstrate that our method outperforms state-of-the-art approaches. In addition, we provide a new perspective on understanding the UDA problem through back-propagation analysis.
% Additionally, we provide critical insights into understanding point cloud UDA problems from the perspective of back-propagation and gradient surgery.

\end{abstract}


\IEEEpeerreviewmaketitle

\section{Introduction}

The point cloud emerges as a crucial data type in real-world 3D media interpretation applications such as robotics and self-driven vehicles \cite{qi2017pointnet,qi2017pointnet++,wang2019dynamic}. As it is costly to obtain enough point cloud data with precise manual annotation \cite{dai2017scannet}, many methods rely on synthetic Computer-Aided Design (CAD) object datasets \cite{vishwanath2009modelnet, chang2015shapenet}. However, the performance of these models tends to degrade when applied to real-world scenarios due to significant domain discrepancies, including variations in shape, differences in density, and occlusions from real-world sensors.
Prevailing methods in point cloud Unsupervised Domain Adaptation (UDA) have \textit{adopted the Multi-Task Learning (MTL) paradigm} to bridge domain gaps. This approach involves designing self-supervised tasks shared by both source and target domains, thereby reducing domain discrepancies by encouraging the backbone network to learn effective feature representations across both domains. Various novel self-supervision tasks have been proposed, such as point cloud reconstruction \cite{shen2022domain}, 3D puzzle sorting \cite{alliegro2021joint}, and prediction of structural attributes \cite{liang2022point}.

\begin{figure*}[tb]
    \centering
    \includegraphics[width=0.62\linewidth]{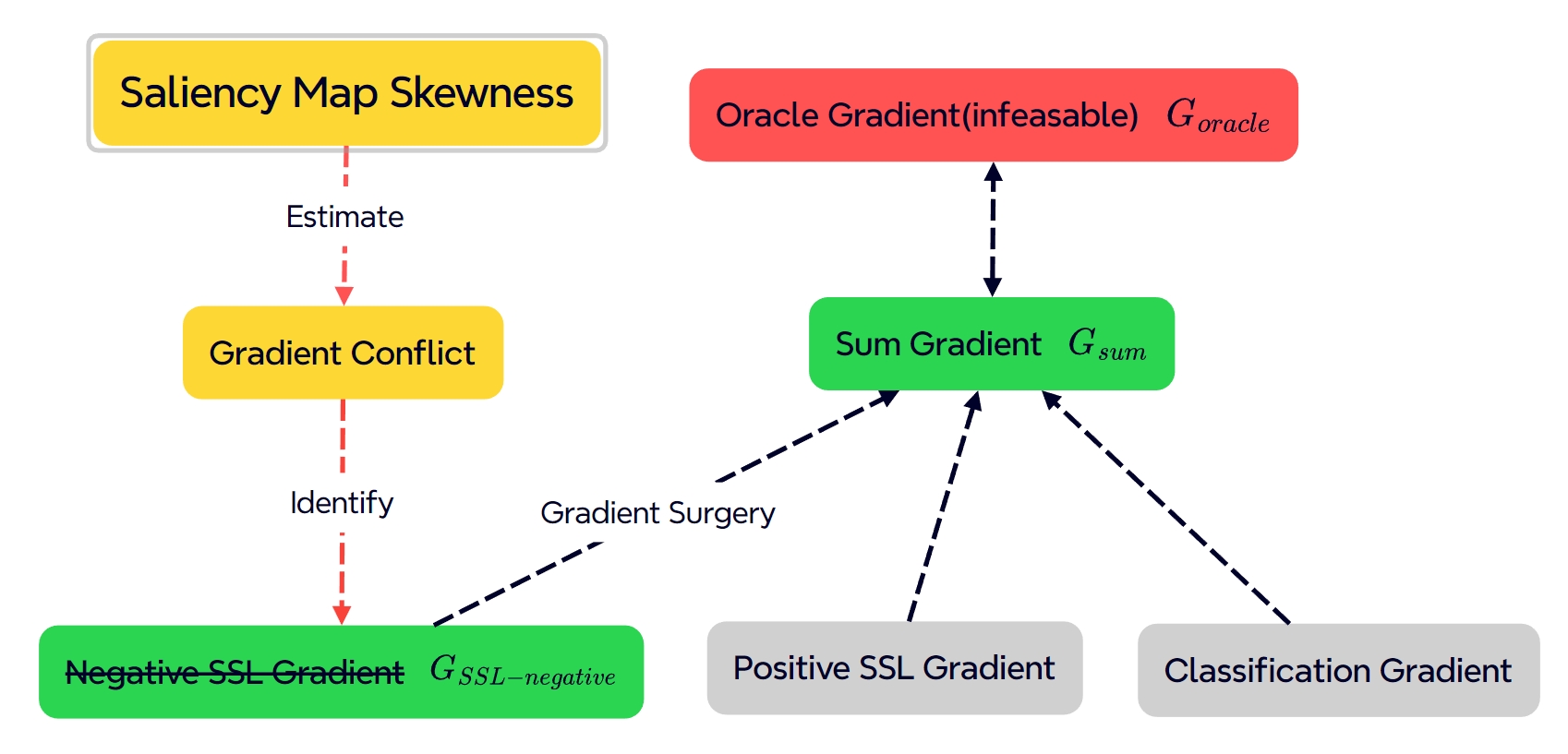}
    \includegraphics[width=0.32\linewidth]{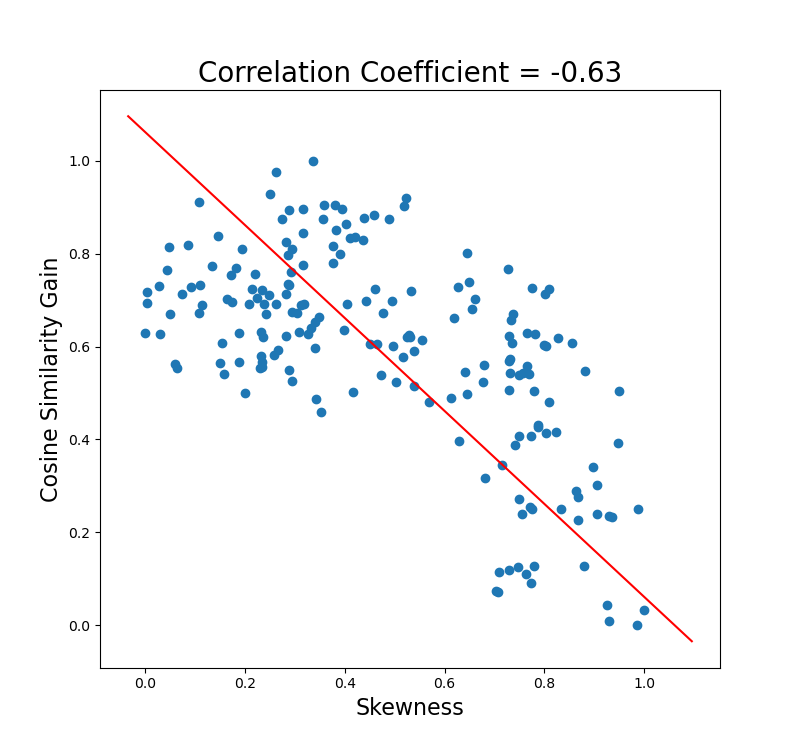}
    
    \caption{The left figure illustrates the design principle of our methods: we first use saliency map skewness to estimate the gradient conflict and then identify negative self-supervision gradient, optimize the sum gradient, making it closer to Oracle gradients. The right figure shows the correlation analysis between the saliency map skewness and gradient conflicts. Each point represents an observation from a specific training step, with its position along the X and Y axes indicating its values for the saliency map skewness and cosine similarity between $G_{SSL}$ and $G_{oracle}$. Lower skewness indicates less conflict and more gain in the cousin similarity. }
    \label{fig:teaser} 
\end{figure*}

\textit{Despite the overall enhancements brought by self-supervision tasks, accuracy on some class degrades compared to methods without adaptation, raising concerns about its effectiveness.} For instance, Zou et.al \cite{zou2021geometry} design a geometry-aware self-training method which improves the classification accuracy of \textit{Chair} from 49.8\% to 74.3\% but decreases \textit{Bathtub} accuracy from 61.5\% to 57.7\%. This observation highlights a significant challenge: why self-supervision tasks lead to such varied class accuracy enhancements and whether we could design a method to relieve negative transfer on certain classes.

To better understand this problem, we analyze it from the perspective of gradient conflicts and the model parameter update process in UDA. As shown in Figure \ref{fig:teaser}, we define the \textit{oracle gradients} $G_{oracle}$ as those induced by the target data and labels, which serve as the oracle for UDA. During the training, the overall gradient $G_{sum}$ consists of two types of gradients contributing to the parameter updates: supervised classification gradients $G_{Cls}$ from the source domain labels, and self-supervised gradients $G_{SSL}$ from auxiliary tasks on both the source and target domains.
As pointed out by Yu et al. \cite{yu2020gradient}, detrimental gradient interference can occur in multi-task learning and hinder optimization progress. Under this setting, by integrating $G_{SSL}$, $G_{sum}$ are expected to be adjusted to better approximate $G_{oracle}$, thus mitigating overfitting to the source domain and promoting effective knowledge transfer \cite{he2022metabalance}.

However, a critical question remains: \textbf{Are all the training samples contributing the beneficial $G_{SSL}$, and should $G_{SSL}$ all be applied uniformly across the model training process?} 
To explore this question, we first conducted a pilot experiment, \textit{Random Gradient Freezing}, where we randomly froze the self-supervision gradients for 50\% of the training samples in each batch. Contrary to our expectations, this did not degrade classification accuracy; in some cases, it even enhanced performance compared to the original method. These findings suggest the presence of gradient conflict, where some conflicting gradients from self-supervised tasks interfere with the primary learning objective, i.e. classification of the target domain.
If we could identify $G_{SSL-negative}$ and adjust $G_{SSL}$, we could then make $G_{sum}$ more closely align with $G_{oracle}$. \textit{However}, in the absence of target domain labels, directly calculating $G_{oracle}$ and quantifying the gradient conflict is infeasible.

To overcome this, \textbf{we design a novel metric based on the skewness of the saliency map to implicitly estimate the gradient conflict thus identifying $G_{SSL-negative}$.}  
The saliency map assigns a score to each point in the point cloud, reflecting its importance in the model's classification decision, and skewness measures the asymmetry in these scores. Our approach is motivated by two key observations. \textit{First, from a quantitative perspective}, our extensive experiments have demonstrated a strong and domain-agnostic correlation between the skewness of the saliency map and the gradient conflict (as shown in Fig. \ref{fig:teaser}). Lower skewness consistently indicates less conflict and more gain in cosine similarity across different domains. \textit{Second, from a qualitative point of view,} high skewness implies that the model heavily relies on a few critical points, which is closely related to the shortcut learning problem \cite{geirhos2020shortcut}. This makes the model more susceptible to performance drops when faced with domain shifts where such points are no longer reliable. Conversely, low skewness indicates that the model's attention is more evenly distributed, enabling it to learn more robust and transferable features. 

Building upon these insights, we propose a novel Saliency Map-based Data Sampling Block (SM-DSB) as a good approximation to the effect of self-supervision gradients for UDA setup and to improve the adaptation performance by filtering out samples with large estimated gradient conflicts. 
The proposed SM-DSB contains two main components: a Measurer module and a Selector module. The Measurer computes the skewness of instance-level saliency maps to estimate the gradient conflict for each sample, based on the observation that higher skewness correlates with larger gradient conflicts (as shown in Fig. \ref{fig:teaser}). The Selector then uses these skewness scores to adaptively select the optimal samples for participation in self-supervised tasks. By filtering out target samples likely to cause negative transfer, the Selector coordinates source and target data to minimize conflicts during joint training.

Our SM-DSB has simple architecture designs and is able to be plugged into most of the mainstream point cloud UDA methods. Our contributions are summarized as follows:
\begin{itemize}
    \item We introduce the skewness of the point cloud saliency map to estimate gradient conflicts in the UDA setting. This helps identify and select samples to mitigate the negative influences of self-supervised gradients.
    \item Our method can be integrated into most mainstream point cloud DA models with little computational overhead. When combined with \cite{cardace2023self}, it achieves state-of-the-art (SOTA) results.
    \item We offer a new perspective on understanding the UDA problem through back-propagation analysis. This encourages further exploration into developing metrics to estimate and address gradient conflicts.
\end{itemize}

% \chensays{GENERAL for Intro: your reader is a reviewer, but he may not get what you want to say at the first minute. Even if a reviewer knows exactly what you have done, he may not remember all the details of the method (not to say he may not have time to catch the method part when reading the intro). INTRO is a summarized paper and is also a teaser of the whole paper. The teaser should be interesting and specifically designed for readers. }

\section{Related Work}
\subsection{Self-supervised 3D Domain Adaptation}
Self-supervised learning (SSL) is a promising research direction in the field of 3D point cloud domain adaptation, as it can generate supervision signals from raw point cloud data, and learn the representation which benefits downstream tasks. Several efficient self-supervised learning models have been proposed recently to capture the domain-invariant geometric features.

Achituve et al. \cite{achituve2021self} propose to deform and reconstruct a region of the shape and analysis the performance of three types of region selection methods.
Zou et al. \cite{zou_geometry-aware_2021} propose the GAST method which learns a shared representation through a novel curvature-aware distortion location prediction for local geometry and rotation angle prediction on mixed-up point clouds to capture the global feature.
Meanwhile, \cite{fan_self-supervised_2022} proposes to predict the scale of the scaled up/down point clouds for global structure modeling, and then reconstruct 3D local regions from the projected 2D planes for local.
\cite{avidan_point_2022} propose to predict masked local structure via estimating point cardinality, position, and normal. 
\cite{cardace2023self} uses a self-distillation paradigm to train a student encoder to match the output of a teacher encode. Two encoders take weakly and strongly augmented point clouds respectively.

However, all of these self-supervision tasks are jointly trained with the classification tasks, they treat all the samples equally and ignore the analysis of which samples are able to bring benefits or negative effects to the two tasks.
Our method designs a novel sample selection block that could be plugged into any of these models to decide which samples to participate in the self-supervision tasks.

\subsection{Data Sampling}
Data sampling methods are proposed to address the problem of imbalanced and noisy labels of datasets that obstruct the training process\cite{wan_survey_2022}. They adjust the frequency and order of training data to facilitate model convergence and improved performance. The major approaches include data re-sampling and loss re-weighting\cite{olvera-lopez_review_2010}.

Data re-sampling techniques include under-sampling and over-sampling \cite{drummond2003c4}. They balance class distribution by excluding data from the majority classes or re-sampling the data from the minority class. 
An instance-level sampling method was designed to mitigate class imbalance for instance segmentation \cite{sun2020crssc}. Zang et al. \cite{zhang2021distribution} proposed using the classification loss on a balanced meta-validation set to adjust feature sampling rates for different classes. 
In addition to hard sampling that removes data (W = 0) or selects data (W = 1) in a meaningful order for training, loss re-weighting addresses the problem above through soft sampling, i.e., adjusting each data loss value weight to optimize the model training. 
For example, \cite{liu2015classification} designed assigns smaller weights to examples with noisy labels and larger weights to clean examples. Zhang et al. \cite{zhang2021dualgraph} proposed re-weighting examples based on the structural relationships between labels to eliminate abnormally noisy examples.  

 To ease the gradient conflict in the MTL framework,  Method "PCGrad" is designed to mitigate gradient differences via "gradient surgery"\cite{yu2020gradient}. Chai et al. 
 \cite{liu2021conflict} introduce Conflict-Averse Gradient descent (CAGrad) which minimizes the average loss function, while optimizes the minimum decrease rate of any specific task’s loss.
 \cite{chai2022model}  proposed a model-agnostic approach to mitigate gradient interference by a gradient clipping rule.

Nonetheless, the aforementioned methods primarily operate within a single domain. Traditional multi-task learning objectives aim to minimize the average loss across all tasks. As such, they do not exhaustively account for the interplay between primary and auxiliary tasks, or between source and target domains. This paper, in contrast, seeks to address these crucial relationships, exploring the dynamics between different tasks and domains, and their influence on the overall learning objective. The goal is to develop an approach to enhance the multi-task learning framework to effectively handle domain adaptation.
% There are also some methods aiming to solve the problem of conflicting gradients. For example, \cite{mansilla2021domain} multi-domain gradient interference in domain generalization.

% \cite{tian2023unsupervised} proposes a novel gradient optimization strategy considering the calibration of dominant and conflicting gradients. But they did not find

We propose to conduct selection with the help of the skewness of the saliency map, a rather \textit{domain-invariant measurer}, which better points the gradients to the direction that is conducive to the convergence of the target domain.

%\IEEEpubidadjcol

\section{Methods}
\begin{figure*}[h]
    \centering
    \includegraphics[width=0.95\linewidth, height=3.2in]{ 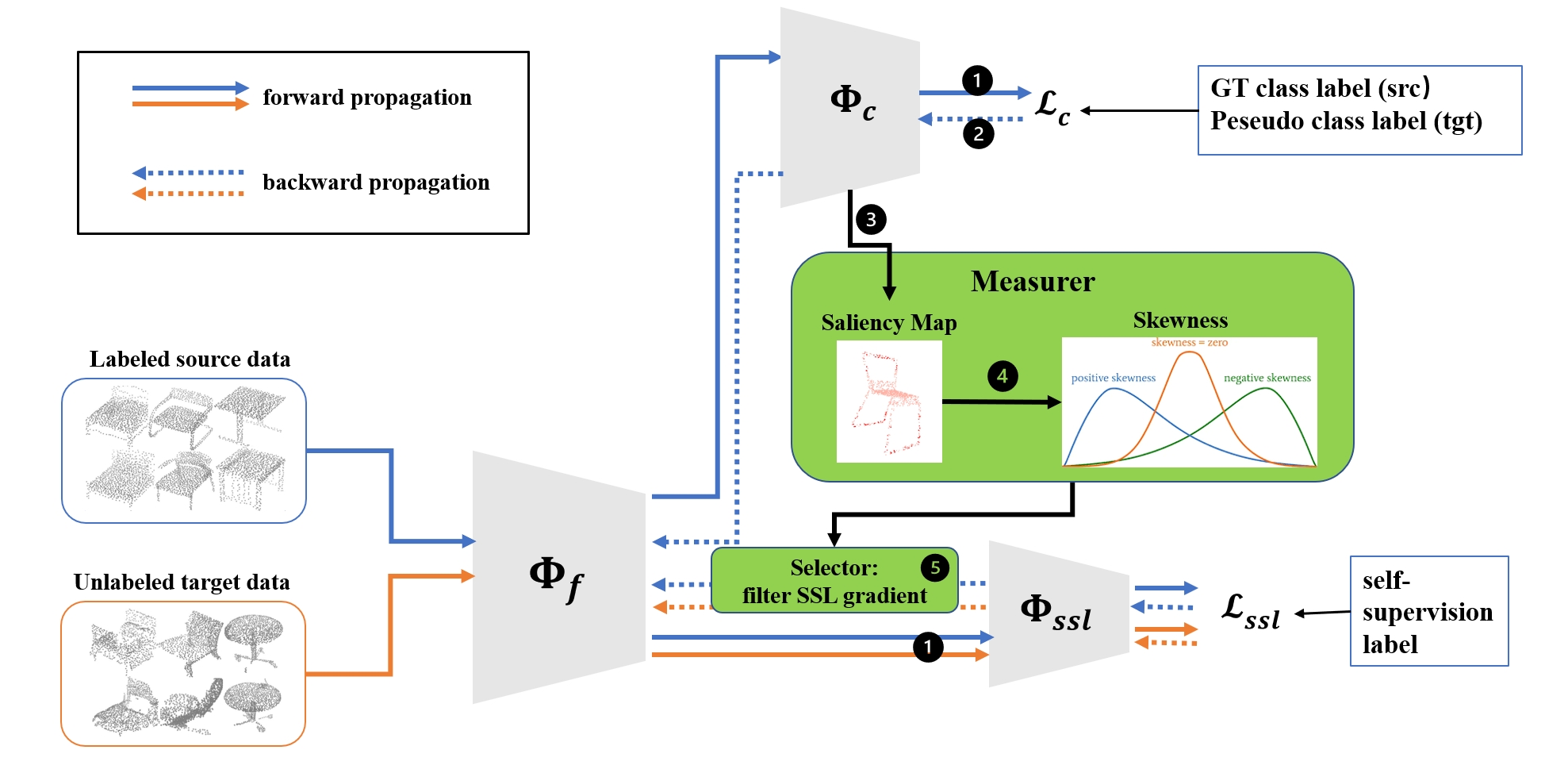}
    \caption{Overview of our SM-DSB plugged into the MTL framework. The measurer computes the skewness of instance-level saliency maps to estimate the gradient conflict for each sample and passes the result to the selector. Then, the selector filters out samples that are suitable for self-supervised training instead of accepting them all.}
    \label{fig:framework}
\end{figure*}

\subsection{Two steps MTL framework}

We study Unsupervised Domain Adaptation (UDA) in point cloud classification, focusing on enhancing the multi-task learning (MTL) framework. Specifically, we have access to a labeled source domain, $S = \{(x_{s}^{i}, y_{s}^{i})\}_{i=1}^{n_{s}}$, and an unlabeled target domain, $T = \{x_{t}^{j}\}_{j=1}^{n_{t}}$, where each point cloud $x \in \mathbb{R}^{N \times 3}$ consists of $N$ coordinates and each label $y \in \mathbb{R}^{K}$ corresponds to $K$-class labels. The objective of UDA is to train a classification model using source domain data that generalizes well to the target domain despite cross-domain discrepancies.

Current point cloud UDA methods follow a two-step approach within a dual-branch MTL framework. In the first step, the model is trained jointly on the labeled source domain for classification and on both source and target domains for self-supervision. The dual-branch framework consists of a \textit{Semantic Classification (SC) branch} and a \textit{Self-Supervised Learning (SSL) branch}. Given the source/target data $x_{s}$/$x_{t}$, features $g_s$/$g_t$ are extracted by the shared encoder $\Phi_{f}$. The classification head $\Phi_{c}$ produces a probability vector $\hat{p} \in [0,1]^{K}$ followed by a softmax function. In the SC branch, the conventional cross-entropy loss $\mathcal{L}_{c}^s$ is utilized to train $\Phi_{f}$ and $\Phi_{c}$ using only the source labels. For the SSL branch, the self-supervision head $\Phi_{ssl}$ and the encoder $\Phi_{f}$ are trained using the loss $\mathcal{L}_{ssl} = \mathcal{L}_{ssl}^s + \mathcal{L}_{ssl}^t$, which depends on (1) any SSL task \cite{achituve2021self, sauder2019self, zou_geometry-aware_2021}; (2) our proposed \textit{SM-DSB} modules to select samples for loss modification. Here, $\mathcal{L}_{ssl}^s$ and $\mathcal{L}_{ssl}^t$ represent the SSL losses on source and target data, respectively.
In the second step, pseudo-labels are generated for the target data by predicting labels based on the model's maximum confidence score per sample. These pseudo-labels are then used alongside the source labels to further fine-tune the model, following the same dual-branch framework. The overall training procedure can be summarized by minimizing the combined loss function:
\begin{equation}    
\mathop{\text{minimize}}\limits_{\Phi_{f},\Phi_{c},\Phi_{ssl}} \mathcal{L}_{c}^s + \mathcal{L}_{ssl}^s + \mathcal{L}_{ssl}^t.
\label{eq:overall-loss}
\end{equation}

\subsection{Overall pipeline}

The SM-DSB method improves the current MTL framework by mitigating potential gradient conflicts. Since target labels are unavailable and oracle gradients cannot be directly computed, finding an appropriate metric to estimate gradient conflict is essential.
As we observed a strong correlation between saliency map skewness and gradient conflicts, we propose using the skewness of the saliency map as an alternative score to quantify and identify sample-level gradient benefits for our learning objective. The SM-DSB method consists of two components: a Measurer that estimates instance-level gradient conflicts by computing saliency map skewness, and a Selector that determines which self-supervision gradients should participate in model updates. 

After obtaining the classification loss $\mathcal{L}_{c}$, the Measurer computes the skewness of instance-level saliency maps. Before backward propagation of the self-supervised loss $\mathcal{L}_{ssl}$, the Selector uses these skewness scores to select optimal samples for self-supervised tasks. This approach ensures that the SM-DSB block effectively enhances UDA task performance by concentrating on beneficial samples from both domains. In the first stage, where no class labels are available, SM-DSB operates exclusively on source domain data. In the second stage, where pseudo labels are generated for the target domain, both source and target domain data are used.

\subsection{SM-DSB}

Following the current mainstream MTL framework in point cloud UDA settings, the training is divided into two steps. During the first stage, we apply self-supervised learning on both domains and supervised learning on just the source domain. For source domain samples $x_{s}^{i}$, after being encoded into feature vectors by feature encoder $\Phi_f$, the feature vectors are first used for the classification task, where cross-entropy functions as the loss function:
\begin{equation}
    \mathop{min}\limits_{\Phi_{f},\Phi_{c}}\mathcal{L}_{c}^{s}=-\frac{1}{n_{s}} \sum_{i=1}^{n_{s}} \sum_{c=1}^CI[c=y_{s}^{i}]\log{x_{s,c}^{i}} ,
    \label{eq:cross-entropy}
\end{equation}
where $x_{s,c}^{i}$ is the $c$-th element of category prediction $x_s^i$ of a source point cloud and $I[\cdot]$ is an indicator function.

The point cloud saliency map \cite{zheng2019pointcloud} is a method used to detect significant points in 3D point clouds. It assigns a score to each point to reflect its contribution to the model-recognition loss, making it easier to identify the critical points in the model. To construct its saliency map, they approximate the contribution of a point by the gradient of classification loss when shifting in the spherical coordinate system. 

To be specific, for a point cloud $x$ consisting of $N$ points $x \triangleq \{p_i\}_{i=1 \dots N}$. When shifting a point $p_i$ towards the center by $\delta$, the classification loss $L$ will increase by $-\frac{\partial L}{\partial r_{i}}\delta$, where $r_{i}$ represents the distance of a point to the spherical core, which is the median of the axis values of all the points in the point cloud. In practice, to allow more flexibility in saliency-map construction, we introduce a parameter $\alpha$ to rescale the point clouds and apply a change-of-variable by $\rho_{i}=r_{i}^{-\alpha} (\alpha>0)$. As a result, the gradient of $L$ can be recalculated by:
\begin{equation}
   -\frac{\partial L}{\partial \rho_{i}} = 
   -\frac{1}{\alpha} \frac{\partial L}{\partial r_{i}}r_{i}^{1+\alpha} 
    \label{eq:saliency-map}
\end{equation}

Finally, we measure the contribution of a point $x_{i}$ by $-\frac{\partial L}{\partial \rho_{i}}$ in the rescaled coordinates. Since $\frac{1}{\alpha}$ is a constant, the saliency score of $p_{i}$ in the saliency map can be considered as:
\begin{equation}
   s_{i}=-\frac{\partial L}{\partial r_{i}}r_{i}^{1+\alpha} 
    \label{eq:saliency-map2}
\end{equation}
where $s_{i}$ is the saliency score of $p_{i}$. For a point cloud sample with a known category label, we can obtain a corresponding saliency map $S \triangleq \{s_i\}_{i=1 \dots N}$. 

During the calculation of the loss, the corresponding saliency map is obtained through an additional gradient backpropagation step mentioned above. The skewness as a measure of the asymmetry of distribution is then computed based on the saliency maps of each point, which could be expressed as Eq. \ref{eq:skewness}, 
\begin{equation}
  sk = \frac{\sum_{i=1}^n (S_i - \bar{S})^3}{n\sigma^3},
  \label{eq:skewness}
\end{equation}
where $\overline{S}$ represents the mean value of its saliency map and $\sigma$ represents its standard deviation. 

The reason for selecting saliency map skewness as an evaluation metric is due to its inherent ability to provide valuable insights into the distribution of point saliency scores. It signifies the presence of 'critical points' in the distribution. In a right-skewed distribution, relatively critical points are fewer but hold more substantial importance. This phenomenon becomes particularly relevant in situations where input disturbances(domain gap) occur and such disturbances are likely to disrupt these critical points, leading to a more pronounced impact on the model's performance. Consequently, skewness in the saliency map effectively represents the model's robust capability to classify samples, serving as a measure of the training stage for the classification task and thus a valuable tool for adjusting task priorities in multi-task learning scenarios. As shown in Figure \ref{fig:teaser}, the skewness of the saliency map exhibits a strong correlation with the complex gradient conflicts. For more details about the skewness of the saliency map, please refer to Section 5.2. 
Note that we apply this method to the target domain where the pseudo labels are leveraged. Experiments show that the skewness score in the target domain, generated by the pseudo labels, is always slightly larger than the correct value due to the presence of more noise compared to the ground truth labels, as shown in Figure \ref{fig:skewness-tgt} (details can be found in Section 5). To avoid mistaking samples with smaller skewness due to noise influence as samples that are already well understood, we apply a small-scale random perturbation to the skewness of all pseudo-labels from the target domain. This is done in the second stage based on the block architecture in the first stage.

Based on the previously computed score set $\{sk^{1}, sk^{2},...,sk^{B}\}$ of all the samples in a mini-batch, we propose to set an adaptive threshold to select useful samples for the self-supervision task. Specifically, we first sort the score set in ascending order and obtain the ranked set $\{sk_{sorted}^{1}, sk_{sorted}^{2}, ..., sk_{sorted}^{B}\}$. Then, we choose $sk_{sorted}^{B\beta}$ as the adaptive threshold $\tau$, where $B\beta$ is rounded to the nearest integer and $\beta$ is the hyperparameter controlling the ratio of the selection. The selection indicates that $\beta$ of the $B$ samples in a mini-batch have larger score values than $\tau$ and only those with high scores will participate in further processing (self-supervision task).

The $\beta$ is set by balancing the sample quality and quantity. More details of the ablation study can be found in the Section V. 
% We conducted ablation studies to analyze how $\beta$ affects the final performance. Experimental results in Section \ref{sec:ana} show that $\beta=0.7$ strikes a good balance between sample quality and quantity, and leads to the best performance. The details can be found in the Section \ref{sec:ana}.
% \xu{ ablation studies should in Sec 4}

The final self-supervision loss functions $\mathcal{L}_{ssl}^s$ combines the data selection procedure in the following:
\begin{equation}
    \mathcal{L}_{ssl}^s = \sum_{b=1}^B\lambda_b\mathcal{L}_{ssl,b}^{'s}  ,
\end{equation}
where
\begin{equation}
	\lambda_b =\left\{
	\begin{aligned}
        &  0 \qquad if \quad sk_b \geq \tau \\ 
	  &  1  \qquad otherwise
	\end{aligned}
	\right.
\end{equation}
Note that the $\lambda_b$ is the indicator denoting whether the $b$-th sample is selected or not and $\mathcal{L}_{ssl,b}^{'s}$ represents the self-supervision loss of the $b$-th sample. 

% Assuming that the loss function of the self-supervision task for each sample is $\mathcal{L}^{'}$, when calculating the overall loss in the batch $\mathcal{L}_{self}^{s}$, unlike the previous method of averaging the loss of all samples, we propose to 
% \begin{equation}
%     \mathcal{L}_{self}^s = \sum_{b=1}^B\lambda_b\mathcal{L}_{self,b}^{'s}
% \end{equation}
% where
% \begin{equation}
% 	\lambda_b =\left\{
% 	\begin{aligned}
%         &  0 \qquad if \quad sk_b \geq \tau \\ 
% 	  &  1  \qquad otherwise. 
% 	\end{aligned}
% 	\right.
% \end{equation}

\begin{algorithm}
\caption{Selection algorithm.}
\label{alg:cap}
\begin{algorithmic}
\Function{select}{X, $\mathcal{L}_{c}$, $\mathcal{L}_{ssl}$, $\beta$}
    \If{X is from S}
        \State $S^* \gets getSaliencyMaps(X, \mathcal{L}_{c,i})$
        \State $score \gets getSkewness(S^*)$
        \State $score_{sort} \gets sort(score)$
        \State $\tau \gets score_{sort}^{B\beta}$
        \For{(i, $x_i$) in X}
            \If{$score_i \geq \tau$}
                \State $\mathcal{L}_{ssl}[i] \gets 0$
            \EndIf
        \EndFor
        \State $\mathcal{L}_{ssl}.backward()$
    \ElsIf{X is from T}
        \State $\mathcal{L}_{ssl}.backward()$
    \EndIf
\EndFunction
\end{algorithmic}
\end{algorithm}

% \noindent\textbf{Second Stage(with pseudo target labels):}
% In the second stage, we use pseudo labels to further guide the model learning. The measurer can now be applied to some of the samples in the target samples.
% Experiments show that 
% Experiments prove that our selection method is robust and can achieve high accuracy on target domain samples and pseudo-labels, and through our design, the accuracy rate is further increased.

\section{Experiment and Results}
\subsection{Dataset and Settings}
\noindent\textbf{Datasets.} For domain adaptation for classification, we use the PointDA-10 dataset, which includes object point clouds from 10 shared classes obtained from three popular point cloud classification datasets: ShapeNet \cite{chang2015shapenet}, ModelNet40 \cite{wu20153d}, and ScanNet \cite{dai2017scannet}. This allowed us to define six different scenarios involving synthetic-to-synthetic, synthetic-to-real, and real-to-synthetic adaptation. The ModelNet-10 (M) dataset consists of 4,183 training and 856 testing point clouds, sampled from 3D CAD models by following the method proposed by \cite{qi2017pointnet++}. Similarly, ShapeNet-10 features synthetic data only. However, ShapeNet-10 is more heterogeneous than ModelNet-10, as it includes more object instances with a larger structure variance. The ScanNet-10 dataset contains 6,110 and 1,769 training and testing point clouds, respectively, collected from partially visible object point clouds of ScanNet, within manually annotated bounding boxes. As the point clouds in ScanNet-10 are obtained from multiple real RGB-D scans, they are often incomplete due to occlusion with contextual objects in the scenes, in addition to realistic sensor noises such as errors in the registration process. Furthermore, we down-sampled a point subset containing 1,024 points from the original 2,048 point clouds provided by PointDA-10. 

Other than the classification task, we further extend our block into the segmentation tasks evaluated on the PointSegDA\cite{achituve2021self} dataset.
PointSegDA comprises four subsets: ADOBE (A), FAUST (F), MIT (M), and SCAPE (S), all based on a human mesh model dataset proposed by \cite{maron2017convolutional}. Each subset, sampled to 2048 points and represents eight shared classes of human body parts. Despite sharing the same classes, the subsets differ in point distribution, pose, and scanned humans, contributing to their uniqueness. PointSegDA differs from PointDA-10 in the
type of the domain shifts and the actual shapes representing deformable objects.
Thus, PointSegDA allows us to evaluate the proposed method in a fundamentally
different setup.

\noindent\textbf{Implementation Details}
The hyper-parameter of $\alpha$ is set to 1, the same as the \cite{zheng2019pointcloud}. 
The mask percentage of $topIndex$ is set to 0.7 empirically, we will later discuss the effect of other percentage options in Section 5.
The batch size of the model is 64 when using DGCNN as backbone network, and 128 for PointNet\cite{qi2017pointnet}.
We keep the other settings such as optimizer, learning rate, and data split strategy the same as \cite{cardace_self-distillation_2023} except for the stop strategy. During the experiment, we found that due to the large difference in dataset scale, adopting a common number of epochs in different adaptation directions would cause obvious overfitting. Therefore, we determined the number of training of the model according to the number of steps. 
In this article, all models are trained on one NVIDIA 3090 GPU while the DGCNN-based network takes 40,000 steps to stop and the PointNet-based as the backbone network takes 10,000 steps to stop.

\begin{table*}[t]
    \centering
    \begin{tabular}{c|cccccccc}
    \hline
    \hline
    Method & $ M \to S$ & $ M\to S^*$ & $S \to M$ & $S \to S^*$ & $S^* \to $M & $ S^* \to S$  & Avg. \\
 \hline
No Adaptation & 83.3 & 43.8 & 75.5 & 42.5 & 63.8 & 64.2 & 62.2\\
 \hline
 PointDAN \cite{qin2019pointdan} & 83.9 & 44.8 & 63.3 & 45.7 & 43.6 & 56.4 & 56.3\\
 \hline
 ImplicitPCDA \cite{shen2022domain} & \textbf{86.2} & 58.6 & 81.4 & 56.9 & 81.5 & 74.4 & 73.2\\
 \hline
 DefRec+PCM \cite{achituve2021self} & 81.7 & 51.8 & 78.6 & 54.5 & 73.7 & 71.1 & 68.6\\
 \hline
 DFAN \cite{shi2022dfan}& 83.7 & 60.2 & 84.0 & 60.3 & 75.3 & 76.3 & 73.3\\
 \hline
 MLSP \cite{liang2022point} & 83.7 & 55.4 & 77.1& 55.6 & 78.2 & 76.1& 71.0 \\
  \hline
GAST\cite{zou2021geometry} & 79.7 & 54.4& 68.0 & 53.4 & 62.9 & 66.7 & 64.2\\
\hline
 Self-dist GCN \cite{cardace2023self} & 83.9 & 61.1 & 80.3 & 58.9 & 85.5 & 80.9 & 75.1\\
 \hline
 \hline
 (GAST+Ours) & 80.8 & 54.9 & 68.7 & 53.1 & 61.3 & 68.0 & 64.5\\
\hline
  (Self-dist GCN+Ours) & 83.6 & \textbf{63.0} & \textbf{84.9} & \textbf{59.5} & \textbf{90.5} & \textbf{81.7} & \textbf{77.2} \\
  \hline
     \hline
    \end{tabular}
    \caption{ Comparative evaluation in classification accuracy (\%) on the PointDA-10 dataset with DGCNN.  Best result on each column is in bold. Same as the \cite{cardace2023self}, \dag Denotes a more powerful variant of DGCNN and results are obtained by performing checkpoint selection on the test set.}
    \label{tab:classification_result}
\end{table*}

\subsection{Comparison with the State-of-the-art Methods}
To show the effectiveness of our method, we plug our SM-DSB into the current best methods\cite{cardace_self-distillation_2023} and compare it against itself and other methods for point cloud classification DA such as \cite{qin2019pointdan, zou2021geometry, liang2022point}, using DGCNN\cite{wang2019dynamic}  and PointNet as our feature extractors. Furthermore, we compare it with a baseline model trained only on the source domain without any adaptation and an oracle model assuming having access to all target data.  It constitutes the lower bound and the upper bound respectively.

As shown in Table \ref{tab:classification_result}, with our proposed block plugged in the Self-dist-GCN methods, it achieves the best performance on 5 of 6 tasks, which fully demonstrates its advantages and the point cloud domain adaptation tasks. In particular, the method improves over the baseline Self-dist-GCN by 1.9\% and 1.4\% on the most challenging synthetic-to-real transfer tasks M→S* and S→S* respectively, highlighting its effectiveness in learning transferable feature representations. 
Compared with other state-of-the-art methods such as DFAN and MLSP, our method yields over 2\% performance gain on average, manifesting its advantage in learning discriminative and transferable feature representations. Moreover, the highest accuracy of \textbf{90.5\%} on S*→M verifies the superiority of the method in reversing the domain transfer direction and transferring knowledge from the real domain to the synthetic domain. 
It can be said that the proposed Self-dist GCN+Ours method has unique insights and advantages in modeling the combination of point cloud geometry and semantics. Its comprehensive performance advantages and state-of-the-art results on related tasks demonstrate the effectiveness and superiority of the method, providing a valuable attempt at transfer learning modeling that focuses on the joint transfer of point cloud geometry and semantics for subsequent point cloud cross-domain learning research. 

On PointSegDA, we tested the DefRec+PCM method, the results are shown in Table \ref{tab:seg_result}. As for MLSP, we tested on PointSegDA when transferring from FAUST to three other domains and the test mean IoU improves from 4\%, 4.5\%, and 4.8\%  when transferring from FAUST to ADOBE, FAUST to MIT, FAUST to SCAPE respectively. The results show the generalization ability of our sampling strategy. Note that the results of DefRec in our table do not fully match the original paper, because their method did not use self-supervision on the source domain and only applied it to the target domain. We have modified it to apply self-supervision to both domains in order to fit the framework. 

\begin{table*}
    \centering
    \small  % 设置字体大小为 \small
    \begin{tabular}{p{1cm}|c|c|c|c|c|c|c|c|c|c|c|c|c}
       \hline
       \hline
       Method  & F\newline$\to$A & F$\to$M &  F$\to$S & M$\to$A & M$\to$F & M$\to$S & A$\to$F & A$\to$M & A$\to$S & S$\to$A & S$\to$F & S$\to$M & Avg.\\
       \hline
       Oracle  & 80.9 & 81.8 & 82.4 & 80.9 & 84.0 & 82.4 & 84.0 & 81.8 & 82.4 & 80.9 & 84.0 & 81.8 & 82.3 \\
        \hline
       w/o  & 78.5 & 60.9 & 66.5 & 26.6 & 33.6 & 69.9 & 38.5 & 31.2 & 30.0 & 74.1 & 68.4 & 64.5 & 53.6 \\
       \hline
       DefRec & 78.4 & 60.7 & 63.5 & 67.4 & 36.2 & 70.1 & 41.6 & 29.8 & 32.1 & 66.1 & 72.8 & 66.4 & 58.1 \\
        \hline
        \hline
       DefRec+\newline
       ours  & \textbf{81.3} & \textbf{61.2} & \textbf{64.3} & \textbf{67.8} & \textbf{36.3}& \textbf{71.6} & \textbf{42.1} & \textbf{30.0} & 31.4 & \textbf{73.2} & 72.2 & \textbf{67.2} & \textbf{67.2} \\
       \hline
       \hline
    \end{tabular}
    \caption{
    % \chensays{where is the iou-average? which one is our method? we can add another double lines to separate others and ours.It works for all tables}
    Test mean IoU on PointSegDA dataset when plugging our SM-DSB block into the Def-Rec methods. F, M, A, and S represent FAUST, MIT, ADOBE, and SCAPE respectively. Our block achieves better performance in 10 out of 12 transfer directions. }
    \label{tab:seg_result}
\end{table*}

\subsection{Results on other Backbones, Models}
In order to evaluate the impact of our method on various backbone networks, we present experimental results obtained by integrating our sampling block into an alternate feature extractor, PointNet. It is important to note that in order to elucidate the detailed mechanisms and isolate the effects of self-training, all experiments are conducted at the initial training phase, which does not involve pseudo labels in the target domain.
For these experiments, we adopt all original settings with two exceptions: the training steps are adjusted to 20,000 steps, and the percentage is set to 80\% when incorporated into Self-dist GCN (PointNet). 

As exhibited in Table \ref{tab:first_step}, for all the comparative experiments, the performance increases notably on average after adding our sampling block, and most of the transfer directions are improved. For a small number of cases where the performance was not as good as the original method, it may be due to the limitation of hyperparameter settings. The subsequent sections will show how hyperparameters could have obvious effects on the final performance in some directions.

To verify the generalization performance of our selection mechanism  
We further plugged our SM-DSB into other different methods such as GAST and DefRec+PCM both using PointNet as the backbone network. 
For GAST, the accuracies improve from 79.7\% to 80.8\%, and from 54.4\%  to 54.9\% transferring from ModelNet to ScanNet and ModelNet to ShapeNet, following \cite{cardace_self-distillation_2023} to get results by performing checkpoint selection on the test set. As for DefRec+PCM, we gain a 2.1\%(47.1\% to 49.2\%) improvement using ModelNet as the source domain and ScanNet as the target. 

% \chensays{should this part in previous SOTA method comparison or here?}

\begin{figure*}[hbt]
    \centering

    \includegraphics[width=0.49\textwidth, height=0.40\textwidth]{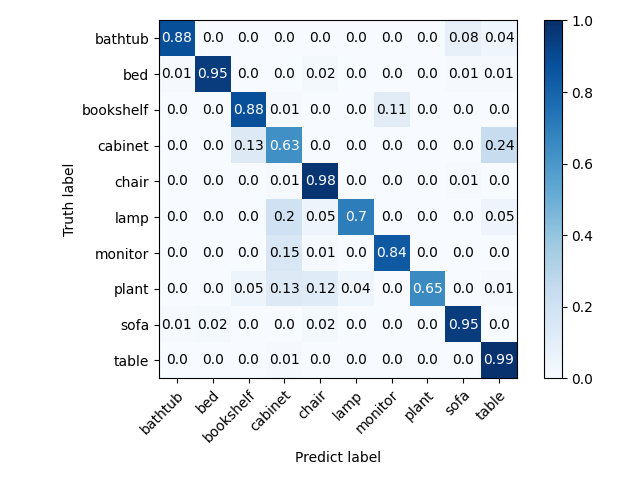}
    \includegraphics[width=0.49\textwidth, height=0.40\textwidth]{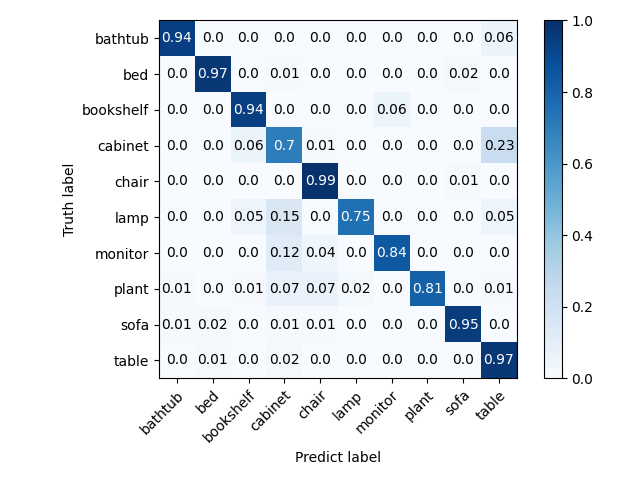}

    \caption{Confusion matrices of classifying testing samples on S* → M, the left and right correspond the methods without and with our SM-DSB under the Self-dist GCN framework.}
    \label{fig:confusion-matrix}
\end{figure*}

\begin{table*}[htb]
 \centering
 \begin{tabular}{c|ccccccc}
 \hline
 \hline
 Method & $ M \to S$ & $ M\to S^*$ & $S \to M$ & $S \to S^*$ & $S^* \to $M & $ S^* \to S$  & Avg. \\
 \hline
 Self-dist GCN & 81.6 & 56.1 & 79.9 & 55.7 & 79.4 & 73.7 & 71.0\\
\hline
 (Self-dist GCN+Ours) & \textbf{84.0} & \textbf{58.5}& \textbf{79.8} & \textbf{56.1}  & \textbf{81.3} & \textbf{74.5} & \textbf{72.4}\\
 \hline
 \hline
 Self-dist GCN(PointNet) & 82.1 & 57.2  & \textbf{77.6} & \textbf{55.0} & 71.0 & 72.1 & 69.2\\
 \hline
 (Self-dist GCN(PointNet)+ours) & \textbf{83.1} & \textbf{57.5} & 77.0 & 54.9 & \textbf{74.3} & \textbf{75.9} & \textbf{70.5} \\
   
% (MLSP+ours) & xx & \textbf{xx} & xx & \textbf{xx} & \textbf{xx} & \textbf{xx} & \textbf{xx} \\
\hline
\hline
\end{tabular}
 \caption{ Plugin effect focusing on just the first step.}
 \label{tab:first_step}
\end{table*}

\section{Analysis}
\subsection{Confusion matrix analysis} 
Confusion matrices in terms of class-wise accuracy are reported in Figure \ref{fig:confusion-matrix} on the task S* → M. Compared with baseline methods, the right one using our data sampling block achieves better performance in most of the classes. Especially the huge improvements for the classes like cabinet and plant that had relatively lower accuracy in baseline methods. Note that an improvement of nearly 20\% is achieved in the plant class.

\subsection{Inconsistency of Gradient Conflicts}
Within the context of multi-task domain adaptation, there exist two distinct types of gradient conflicts. The first type, inter-task gradient conflicts, can be observed during the course of training. The second type is the divergence between the aggregated gradients across tasks and the so-called 'standard' ground truth gradients from the classification label in the target domain. Due to the unavailability of target domain labels, this discrepancy can be evaluated but not directly optimized.

\begin{figure}[htp]
    \centering

    \includegraphics[width=0.24\textwidth]{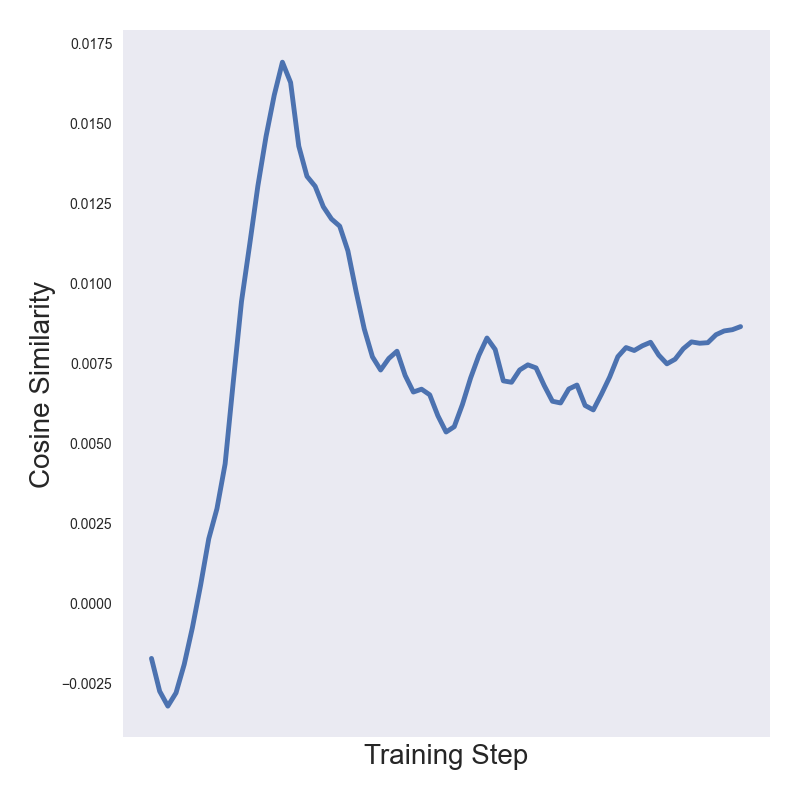}
    \includegraphics[width=0.24\textwidth]{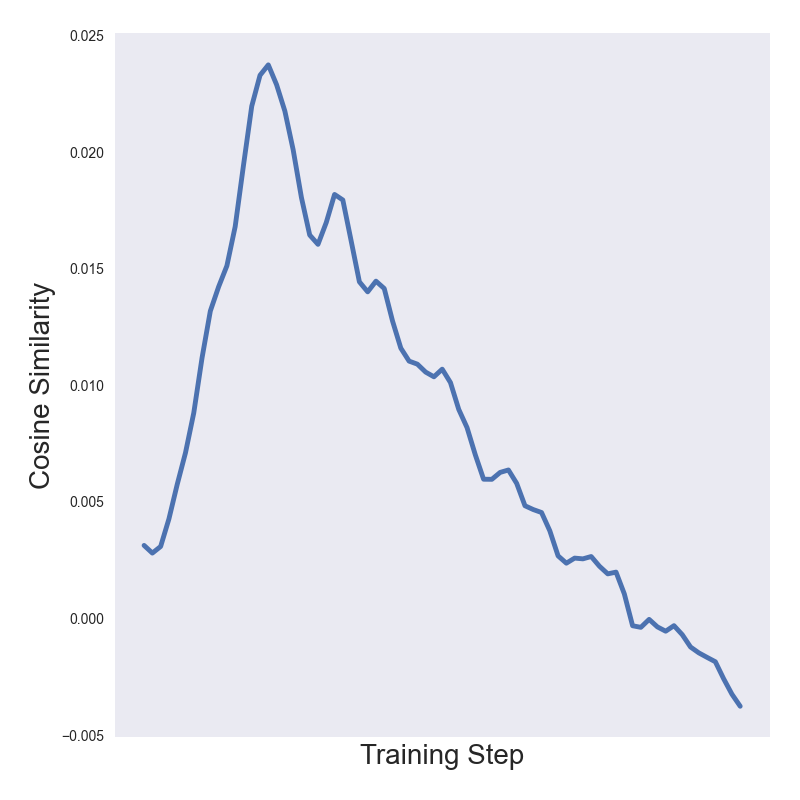}
    % \hfill

    \caption{Conflict inconsistency evaluated on S$\to$M. The figure on the left shows the cosine similarity between aggregated gradients across tasks and the \textit{oracle} gradients. The figure on the right represents the cosine similarity of the inter-task gradient between the self-supervision task and the gradient of the classification task. }
    \label{fig:conflict-inconsistency}
\end{figure}

Our experiments have corroborated the discordance between these two classes of gradients. As depicted in Figure \ref{fig:conflict-inconsistency}, if only inter-task gradient conflicts were accounted for, as is the case in conventional multi-task learning approaches, the similarity remains largely stable throughout the later stages of training. Conversely, the similarity between the aggregated gradients and the ground truth gradients of the target domain diminishes significantly over time, even turning negative at certain points.

These findings imply that a sole focus on inter-task gradient conflicts, as is typical in traditional multi-task learning, can lead a model to converge to solutions that are incompatible with the target domain, culminating in negative transfer. Therefore, a more comprehensive optimization objective, which considers the similarity of both types of gradients, is required to maximize positive transfer in multi-task adaptation scenarios. Despite the 'standard' ground truth gradients being unfeasible during training, the necessity for an alternative, such as the skewness of the saliency map, is underscored by our findings.

\subsection{Insights about saliency map skewness and Gradient Conflict}

\begin{figure}[htp]
    \centering
 
    \includegraphics[width=0.20\textwidth, height=0.20\textwidth]{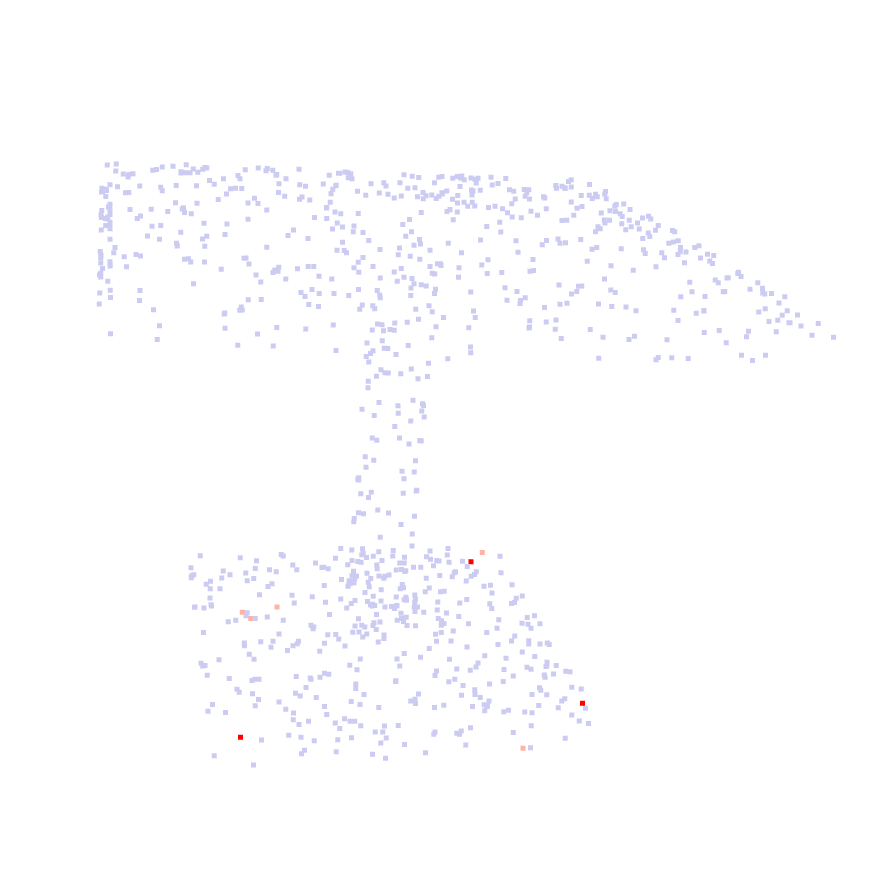}
    \includegraphics[width=0.20\textwidth, height=0.20\textwidth]{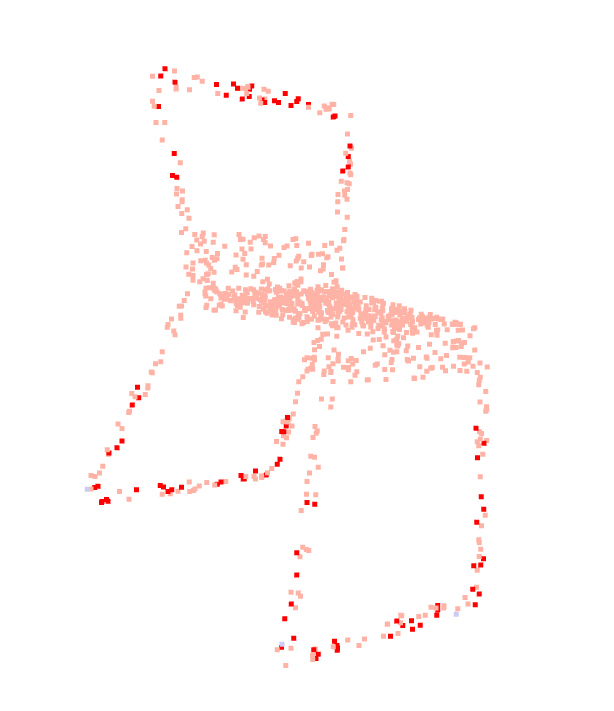}

    \caption{On the left is a table with a high skewness score, while on the right is a chair with a low skewness score. The red region indicates the salient area, whereas the pink and blue regions are less significant. High skewness (right-skewed) signifies a concentration of a few critical points, making the model more sensitive to input disturbances and potentially less robust. Conversely, low skewness (left-skewed) shows more dispersed attention, which indicates the model's ability to maintain classification performance under domain changes.}
    \label{fig:skewness-plot}
\end{figure}

\noindent\textbf{Causal Link Between Skewness and Gradient Conflicts:} In subsequent experiments, we present evidence supporting a causal connection between the skewness and gradient conflicts, as guided by the Additive Noise Model (ANM) propounded in the paper \cite{hoyer2008nonlinear}. Specifically, when skewness is construed as the cause and gradient conflicts as the effect, the ANM fit score is observed to be 0.16. Conversely, when the causal relationship is reversed, the score increases to 0.25. It is noteworthy that a lower score is indicative of a more plausible causal relationship. Further tests were performed to investigate the causation between skewness and conflicts among gradients from the classification task, self-supervision tasks, and the standard gradient independently. The results reveal a more pronounced causal direction toward the classification gradient, implicitly affirming the robustness of our method. This robustness stems from our method's focus on adjusting the priority of the classification task, thereby not necessitating dependence on specific self-supervision tasks. This nuanced approach bolsters the versatility and broad applicability of our method.

\begin{figure}[tp]
    \centering
    \includegraphics[width=0.20\textwidth, height=0.20\textwidth]{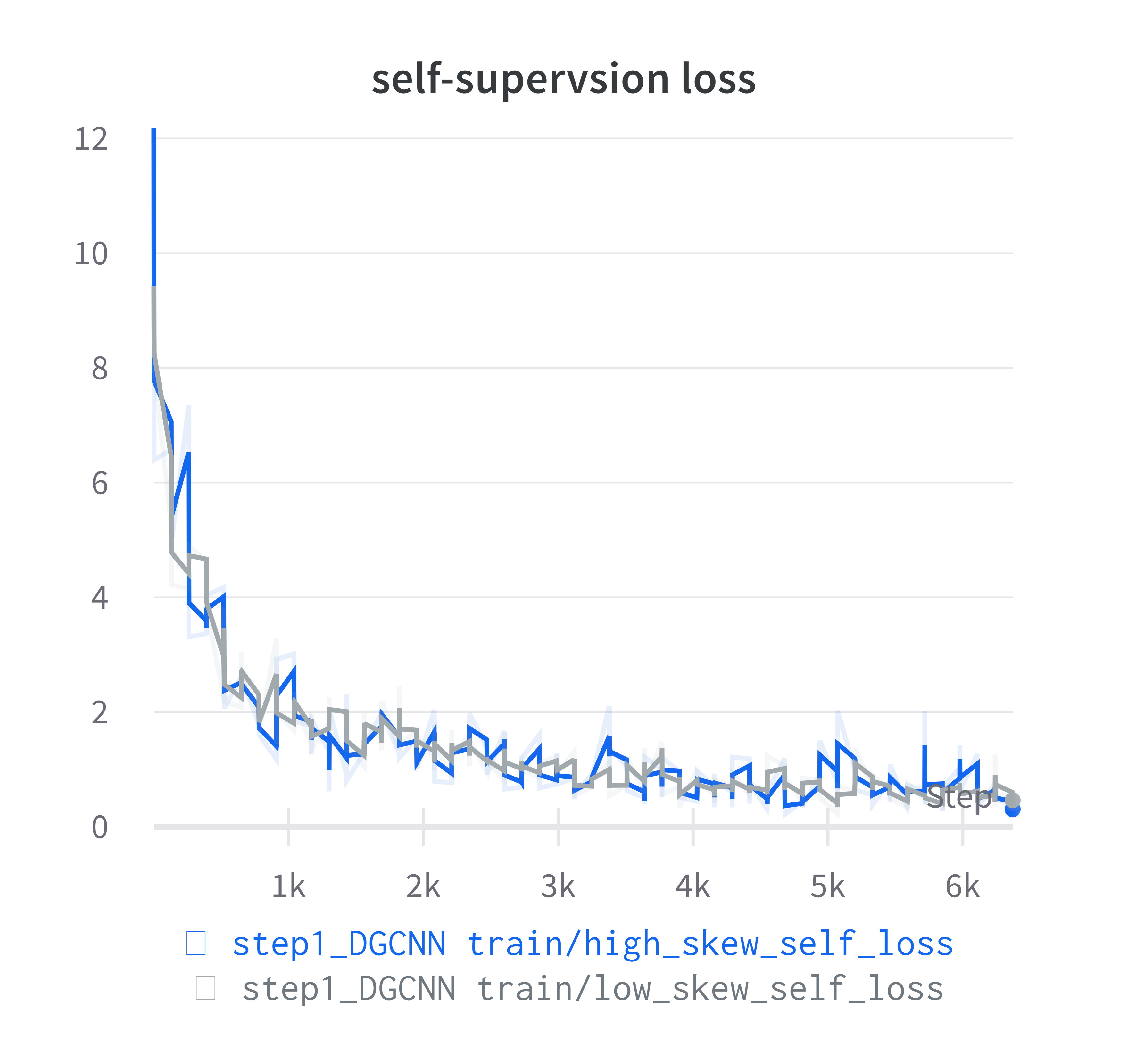}
    \includegraphics[width=0.20\textwidth, height=0.20\textwidth]{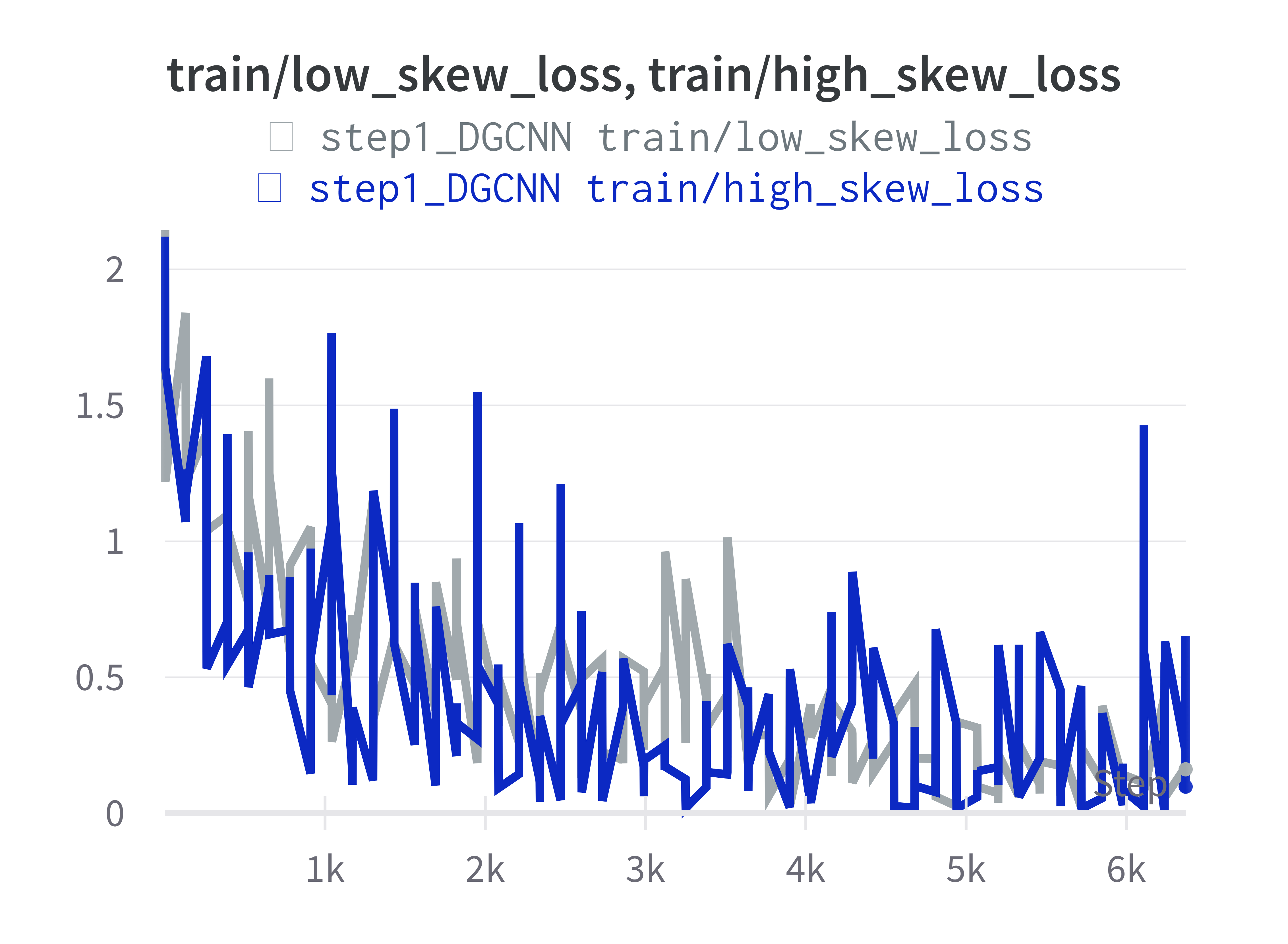}
    \caption{Loss of self-supervision task and classification task. The blue refers to the samples with high skewness while the grey refers to low. The high and low are determined by the adaptive threshold which will be explained in Section 3. }
    \label{fig:unstable-loss}
\end{figure}

\noindent\textbf{Reasons for choosing skewness as evaluating metric:} The saliency map plays a crucial role in assessing the importance of points, as it underscores their contribution to the model's recognition loss. Skewness, being a measure of asymmetry, provides valuable insights into the distribution of point saliency scores. Specifically, positive skewness (right-skewed) signifies that the right side of the distribution tail is fatter than the left side, indicating a smaller number of points with high scores—termed as 'critical points'—in the distribution. It is shown in Figure \ref{fig:skewness-plot} where we adopt the same visualization methods as \cite{liu2023interpreting}. The salient values are divided into three levels ordered by the given interval with [0,$\frac{1}{3}d,\frac{2}{3}d,d$] where $d$ denotes the range of the saliency values. The model successfully captures the disperse-distributed critical points on key edge contours while failing in the left one. In scenarios involving input disturbances, right-skewed distributions are more likely to experience disruption in critical points compared to left-skewed distributions. Consequently, such disturbances are expected to exert a more pronounced impact on the model's performance. On the contrary, the network's attention may be overly dispersed across the whole space. When the input sample is disturbed or the domain changes, the broadly scattered critical points have a lower chance of exceeding the limited perceptual scope of the network. Thus, the network can more easily identify them correctly, avoiding a sharp drop in classification performance.  It can be demonstrated by plotting the loss of samples with different skewness during the training process in Figure \ref{fig:unstable-loss}, the effectiveness of the selection strategy is also verified - samples with larger skewness suffer from more unstable loss (larger variance observed in the training curve). It indicates that the semantic features in the high-skewness group are more ambiguous than the low-skewness group, leading to unstable predictions.  To sum up, saliency map skewness effectively represents the model's robust capability to classify samples at this particular moment, thereby serving as a measure of the training difficulty for the classification task. Consequently, it can be integrated into the adjustment of task priorities in multi-task learning scenarios. 

\noindent\textbf{Emperical Explanation about their Correlation} 
Multitask learning models are sensitive to task priority \cite{kendall2018multi}. Inappropriate task priority may cause a performance decrease, which is especially problematic if a subset of tasks is of primary interest and the others are used only to improve the representation learning.\cite{liu2019loss}
For instance, imbalances in task difficulty can lead to unnecessary emphasis on easier tasks, thus neglecting and slowing progress on difficult tasks. \cite{guo2018dynamic}  
A task weight is commonly defined as the mixing or scaling coefficient used to combine multiple loss objectives.
For our multitask learning in UDA classification, since we could use the saliency map skewness as a metric to evaluate the model's capacity to comprehend the samples' class, which indirectly indicates the task difficulty on the specific sample, we could adjust the balance by enforcing the model to be updated just through the gradient from the more difficult task and not participate in the easier one.  

% and making sure the semantic information is injected into the backbone model

% \subsection{Comparison with other selection methods}

% \noindent In this section, we aim to validate the efficacy of our proposed data selection strategy. 
% Specifically, we conducted a comparative analysis with alternative data selection methodologies, namely: (1) random selection, and (2) selection based on cross-entropy loss, where we selected samples exhibiting the top-K highest losses. To ensure a fair comparison and maintain uniformity, we set the selection ratio $\beta$ in each batch to be equal to that used in our method.
% As illustrated in Figure \ref{fig:other-drop}, our method outperforms the other techniques in terms of overall performance. This superior performance verifies the effectiveness of our data selection strategy, particularly in scenarios where the goal is to minimize losses and enhance accuracy.

% \begin{figure}[htb]
%     \centering
% \includegraphics[width=0.48\textwidth, height=0.35\textwidth]{abalation.png}
%     \caption{Comparison of other selection mechanisms, the yellow and blue line indicates random selection and select the samples that generate the top-K largest loss respectively.}
%     \label{fig:other-drop}
% \end{figure}

\begin{figure}[htp]
    \centering
    \includegraphics[width=0.45\textwidth, height=0.30\textwidth]{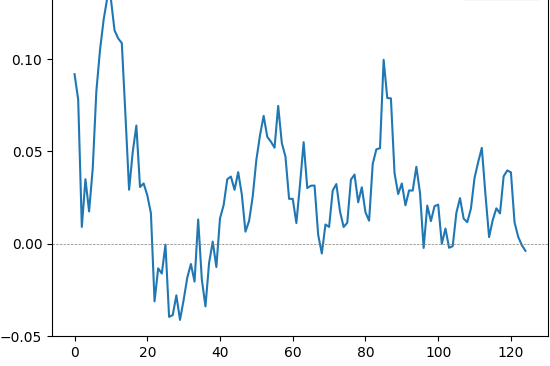}
    \caption{Skewness score discrepancy in the target domain. The x-axis represents the epoch number during training, while the y-axis represents the difference in skewness scores between the noisy pseudo labels and the ground-truth labels.}
    \label{fig:skewness-tgt}
\end{figure}

\subsection{Stability of skewness score in target domain}
During the second stage of the sample selection (section 3.3), the target domain data are incorporated to compute the skewness scores using the pseudo labels obtained after the first stage training. However, the significant noise in these target domain labels may mislead the model by introducing more high-entropy samples (prone to produce misclassification).

To address this, the difference between skewness scores based on the noisy pseudo labels versus the accurate ground truth labels is analyzed to quantify the degree of noise in Figure \ref{fig:skewness-tgt}. A relatively small discrepancy of 0.1 is observed.
Therefore, we manually add a small Gaussian perturbation with a mean of 0.1 to the scores using the pseudo labels, in order to account for the noise effects before calculating the skewness threshold and the scores.

\subsection{Hyper-parameter analysis}
\begin{figure}[htp]
    \centering
        \includegraphics[width=0.24\textwidth, height=0.24\textwidth]{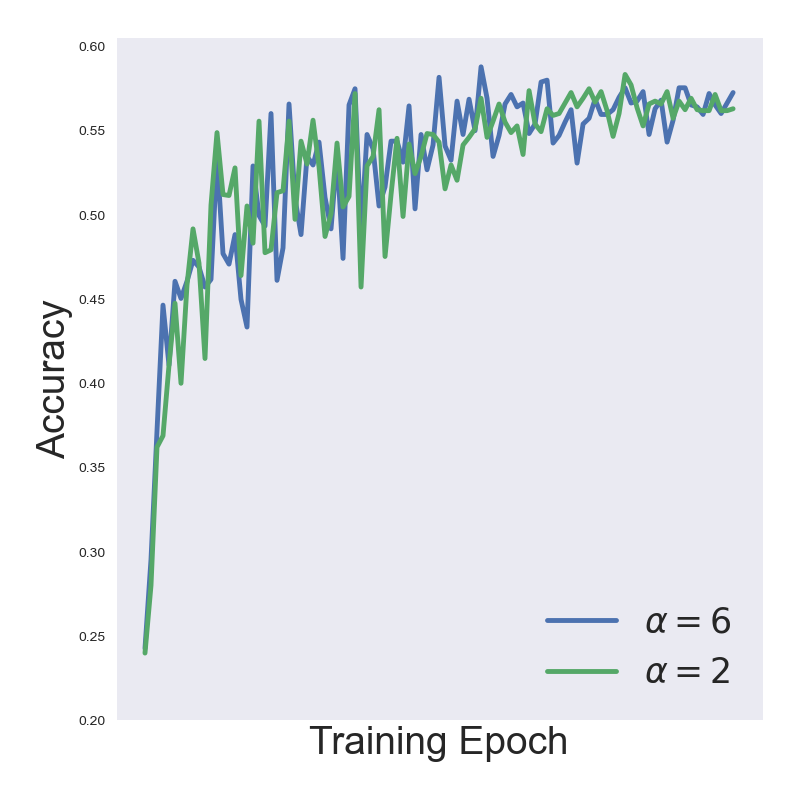}
        \includegraphics[width=0.24\textwidth, height=0.24\textwidth]{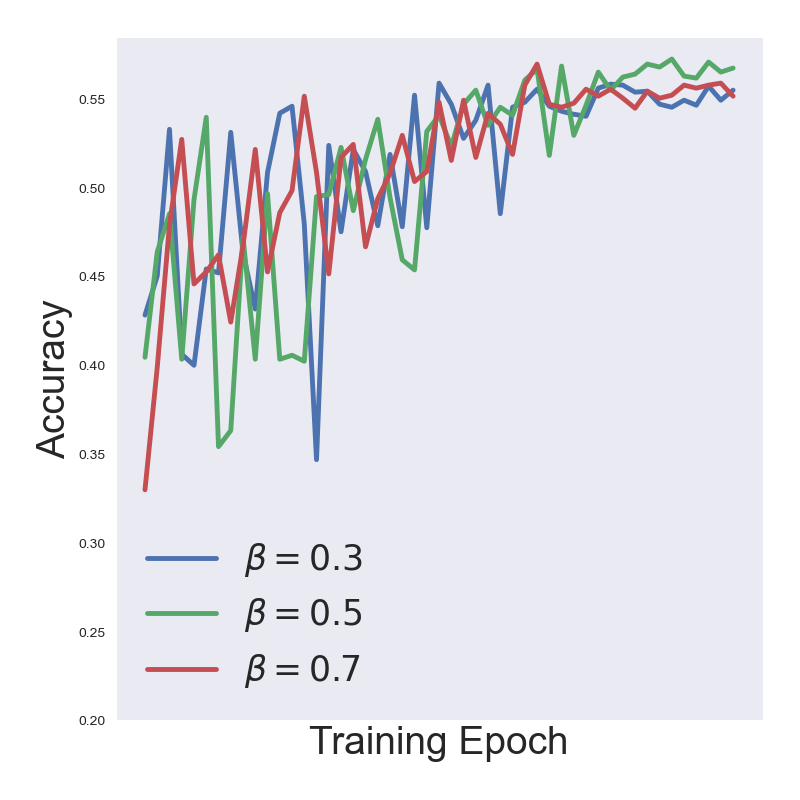}
    \caption{The target accuracy under different hyper-parameter settings evaluated on M$\to$$S^*$. The left figure is about the scaling factor $\alpha$ while the right is about the ratio factor $\beta$.}
    \label{fig:other-hyper-parameter}
\end{figure}

We also evaluate our methods with different hyper-parameters, and the results are visualized in Figure \ref{fig:other-hyper-parameter}. Almost no differences can be observed when selecting different scaling factors $\alpha$. 

\begin{figure}[htp]
    \centering
        \includegraphics[width=0.24\textwidth]{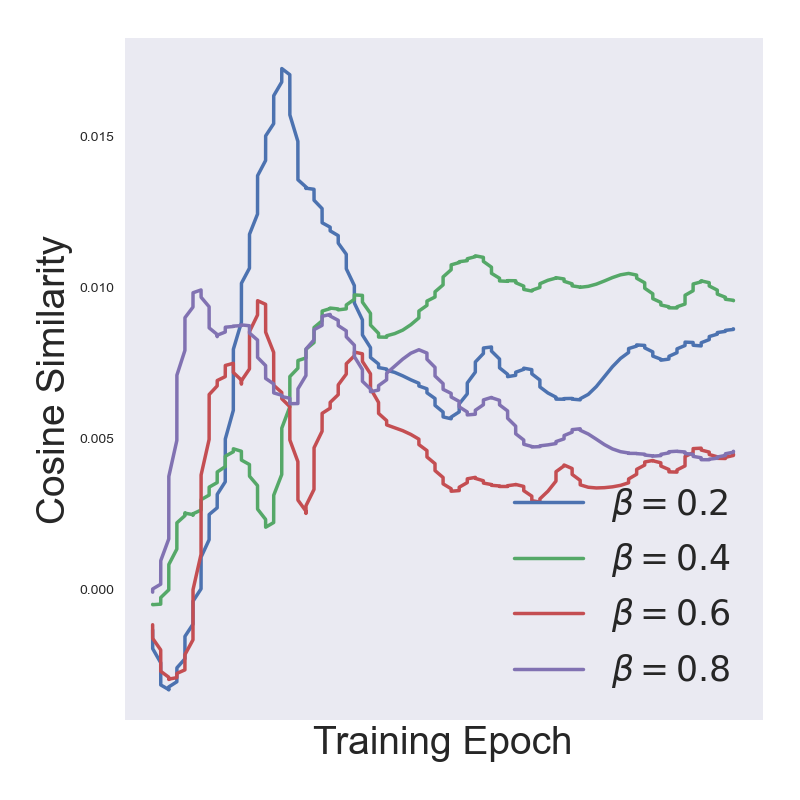}
        \includegraphics[width=0.24\textwidth]{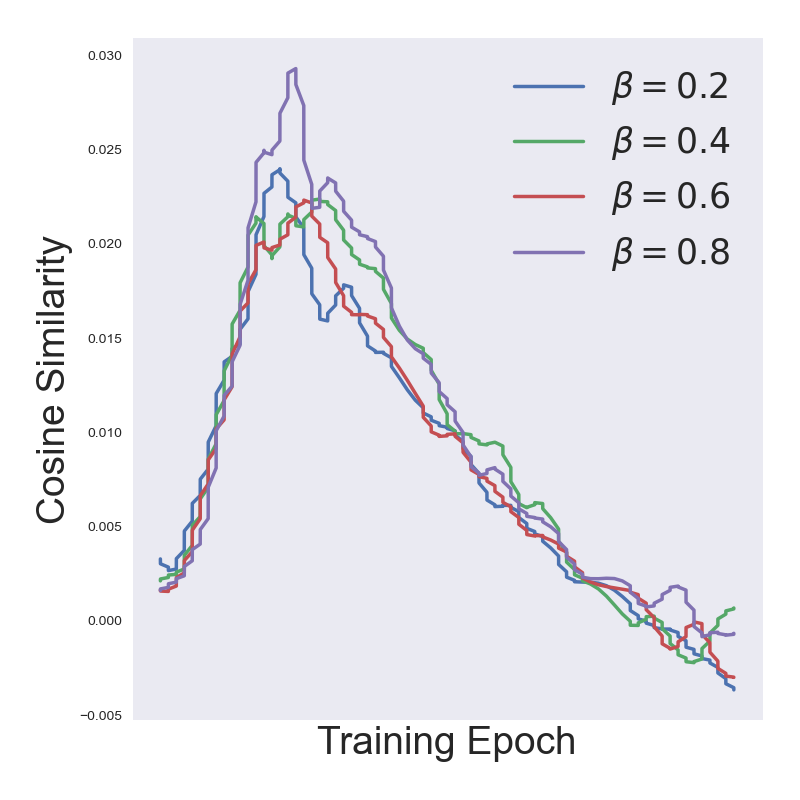}
    \caption{Conflict inconsistency evaluated on S$\to$M. The figure on the left shows the cosine similarity between the gradient sum of the self-supervision task and classification task and the standard gradient. The figure on the right represents the cosine similarity between the gradient of the self-supervision task and the gradient of the classification task. }
    \label{fig:conflict-inconsistency-multi}
\end{figure}

As for selecting the ratio $\beta$, it is highly relevant to the iteration numbers and the dataset scales. There exist better options in different transfer directions. Choosing the $\beta$ as 0.7 for all directions of migration can yield significant improvements. 
For a more detailed analysis, We conduct experiments to reveal how the selection ratio $\beta$ impacts the training process.
First, the ratio settings $\beta$ and the number of iterations exhibit a strong correlation, leading to divergent performance behaviors at different stages of training, \textit{i.e.} $\beta$ that are ideal at an initial training stage may become unsuitable later on and vice versa, as shown in Figure \ref{fig:conflict-inconsistency-multi}. Therefore, comprehensive evaluations of a model should consider performance behaviors over all stages of training, rather than at isolated time points.

\begin{figure}[htp]
    \centering
        \includegraphics[width=0.24\textwidth]{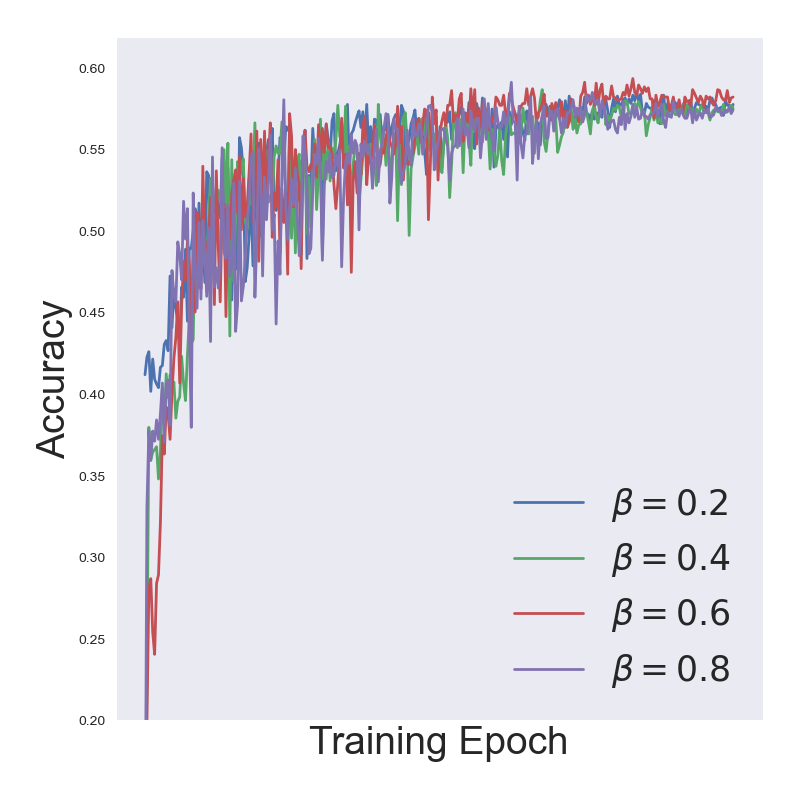}
        \includegraphics[width=0.24\textwidth]{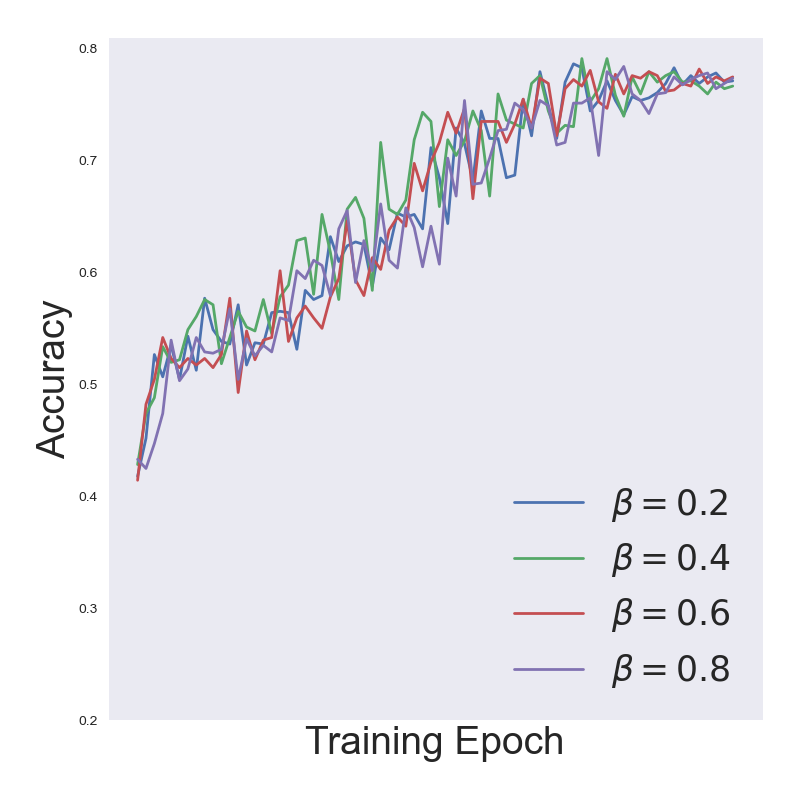}
    \caption{Validation accuracy of different $\beta$ setting evaluated on M$\to S^*$(left) and S$\to$M(right).}
    \label{fig:diff-direction}
\end{figure}

Also, according to results in Figure \ref{fig:diff-direction}, substantial discrepancies were observed in the final results across different transfer directions. For certain directions, varying the ratios led to markedly divergent end performance, whereas for others, the final results converged to largely similar values. A single set of balanced ratios, even if derived from extensive tuning for certain directions, does not necessarily generalize to optimal performance for all directions. Sensitivity analyses of this kind can provide informative insights to guide the pursuit of maximal domain adaptation for diverse applications of multi-task learning.

% \begin{figure}[htp]
%     \centering
%         \includegraphics[width=0.22\textwidth, height=0.20\textwidth]{m2sc_para_former.png}
%         \includegraphics[width=0.22\textwidth, height=0.20\textwidth]{m2sc_para_latter.png}
%     \caption{Validation accuracy of different $\beta$ setting evaluated on M$\to$S during the training process. The figure on the left shows the former part of the training and the right shows the latter. \tjq{- There are some issues with the visualizations presented in the paper. For instance, the font sizes in the titles of Figures 10 and 11 are different.
% - Similarly, Legends in Figures 12 and 13 have too small font sizes, making them hard to read. It
% would be great if the author could use more reader-friendly font families and sizes.}}
%     \label{fig:conflict-inconsistency-multi}
% \end{figure}

\subsection{Quality Analysis}
\begin{figure}[htp]
    \centering
        \includegraphics[width=0.22\textwidth, height=0.20\textwidth]{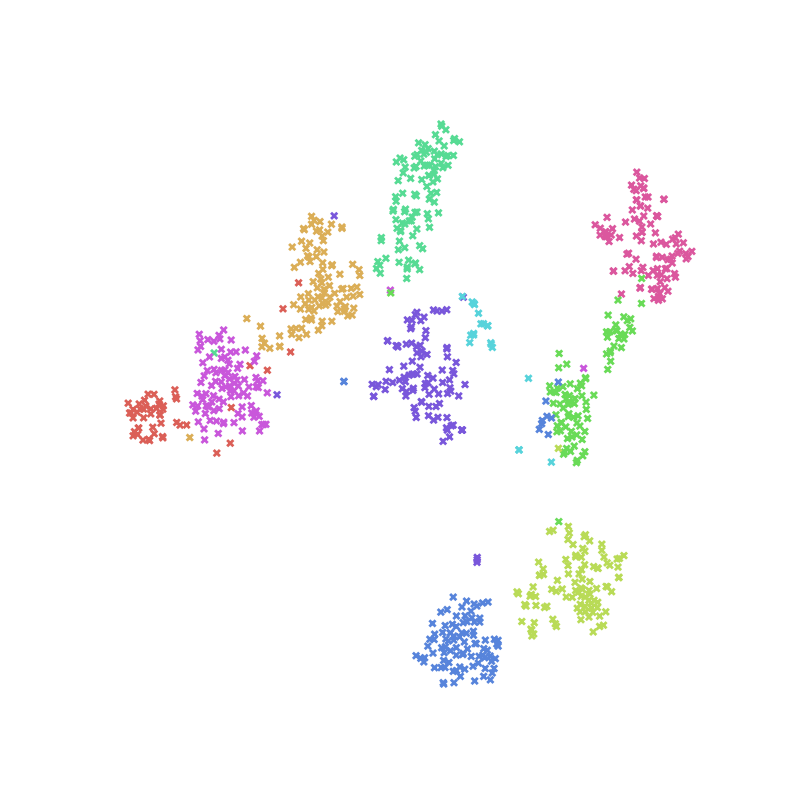}
        \includegraphics[width=0.22\textwidth, height=0.20\textwidth]{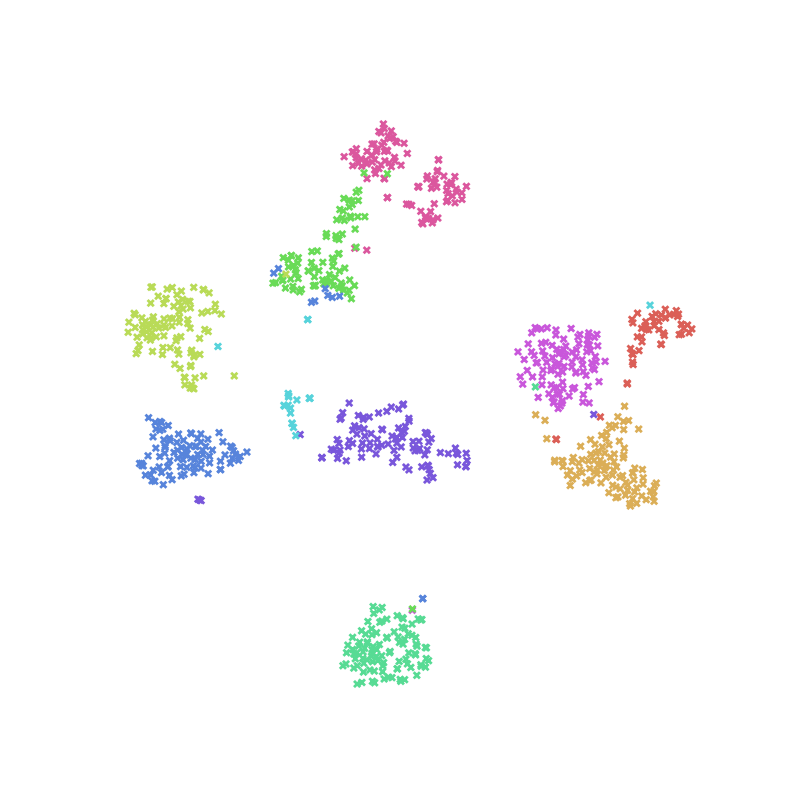}
    \caption{The t-SNE visualization of feature distribution on
the target domain evaluated on $S^*\to$M. Colors indicate different classes.}
    \label{fig:tsne}
\end{figure}

We utilize t-SNE to visualize the feature distribution on the target domain of the UDA task S* → M of the baseline without and with our sampling block in Figure \ref{fig:tsne}. In comparison to the baseline t-SNE visualization, the modified approach yielded cluster boundaries that were substantially more distinct and thus can be more discriminative than those of the baseline. In view of an imbalanced class distribution, the minor class(blue-green color near the center of the figure) exhibits tighter coherence.

\section{Conclusion}
In this paper, we propose a novel sampling block called SM-DSB that uses the skewness of the point cloud saliency map as a metric to estimate gradient conflict while the target classification label is unfeasible in UDA setting. The block can filter samples in each batch to participate in self-supervision tasks based on the metric while introducing few computation complexities. 
Our SM-DSB block could be plugged into most of the mainstream point cloud DA models and improve their performance accordingly, among which the combination with the Self-dist-GCN model results in SOTA on the PointDA dataset.
At present, we only selected a representative third-order statistic in the saliency map of the point cloud as the selection standard. In future work, we will extend to explore other metrics to estimate the gradient conflict, both accurately and fast.\\[0.5cm]

\noindent\textbf{Data Availability} This manuscript develops its method based on the publicly available datasets: PointDA-10\cite{qin2019pointdan} and PointSegDA\cite{achituve2021self}. There is no specific associated data with this manuscript.

\ifCLASSOPTIONcaptionsoff
  \newpage
\fi

\bibliographystyle{IEEEtran}
\bibliography{cite}

\end{document}